\documentclass[11pt]{article}
\usepackage{graphicx} % Required for inserting images
\usepackage{geometry}
\usepackage[ruled, vlined, linesnumbered]{algorithm2e}
\usepackage{listings}
\usepackage{xcolor}
\usepackage{microtype}
\usepackage{float}
\usepackage{xurl} 
\usepackage[colorlinks=true, urlcolor=blue, citecolor=blue]{hyperref} % Optional: Makes links clickable and blue

\colorlet{punct}{red!60!black}
\definecolor{background}{HTML}{EEEEEE}
\definecolor{delim}{RGB}{20,105,176}
\colorlet{numb}{magenta!60!black}

\lstdefinelanguage{json}{
    basicstyle=\normalfont\ttfamily,
    numbers=left,
    numberstyle=\scriptsize,
    stepnumber=1,
    numbersep=8pt,
    showstringspaces=false,
    breaklines=true,
    frame=lines,
    backgroundcolor=\color{background},
    literate=
     *{0}{{{\color{numb}0}}}{1}
      {1}{{{\color{numb}1}}}{1}
      {2}{{{\color{numb}2}}}{1}
      {3}{{{\color{numb}3}}}{1}
      {4}{{{\color{numb}4}}}{1}
      {5}{{{\color{numb}5}}}{1}
      {6}{{{\color{numb}6}}}{1}
      {7}{{{\color{numb}7}}}{1}
      {8}{{{\color{numb}8}}}{1}
      {9}{{{\color{numb}9}}}{1}
      {:}{{{\color{punct}{:}}}}{1}
      {,}{{{\color{punct}{,}}}}{1}
      {\{}{{{\color{delim}{\{}}}}{1}
      {\}}{{{\color{delim}{\}}}}}{1}
      {[}{{{\color{delim}{[}}}}{1}
      {]}{{{\color{delim}{]}}}}{1},
}

\title{Deep Researcher with Sequential Plan Reflection and Candidates Crossover (Deep Researcher Reflect Evolve)}
\author{
  Saurav Prateek \\
}
\date{January 2026}

\begin{document}

\maketitle

\begin{abstract}
This paper introduces a novel \textbf{Deep Researcher} architecture designed to generate detailed research reports on complex PhD-level topics by addressing the inherent limitations of the Parallel Scaling paradigm. Our system utilizes two key innovations: \textbf{Sequential Research Plan Refinement via Reflection} and a \textbf{Candidates Crossover} algorithm.

The sequential refinement process is demonstrated as an efficient method that allows the agent to maintain a centralized \textbf{Global Research Context}, enabling it to look back at current progress, reason about the research plan, and intelligently make changes at runtime. This dynamic adaptation contrasts with parallel approaches, which often suffer from "siloed knowledge". The \textbf{Candidates Crossover} algorithm further enhances search efficiency by deploying multiple LLM candidates with varied parameters to explore a larger search space, with their findings synthesized to curate a comprehensive final research response. The process concludes with \textbf{One Shot Report Generation}, ensuring the final document is informed by a unified narrative and high fact density.

Powered by the \textbf{gemini-2.5-pro} model, our Deep Researcher was evaluated on the \textbf{DeepResearch Bench}, a globally recognized benchmark of 100 doctoral-level research tasks. Our architecture achieved an overall score of \textbf{46.21}, demonstrating superior performance by surpassing leading deep research agents such as \textbf{Claude Researcher} (45) \cite{clauderesearcher}, \textbf{Nvidia AIQ Research Assistant} (40.52) \cite{nvidiaaiqdeepresearch}, \textbf{Perplexity Research} (40.46) \cite{perplexitydeepresearch}, \textbf{Kimi Researcher} (44.64) \cite{kimiresearcher} and \textbf{Grok Deeper Search} (38.22) \cite{grokdeepresearch} present on the DeepResearch Bench’s actively running leaderboard \cite{deepresearchbenchleaderboard}. This performance marginally exceeds our previous work, \textbf{Static-DRA} (34.72) \cite{staticdrapaper}, and reinforces the finding that sequential scaling consistently outperforms the parallel self-consistency paradigm. The entire source code, outputs and benchmark results are open-sourced at \url{https://github.com/SauravP97/deep-researcher-reflect-evolve/}
\end{abstract}

\begin{figure}[H]
    \centering
    \includegraphics[width=0.9\linewidth]{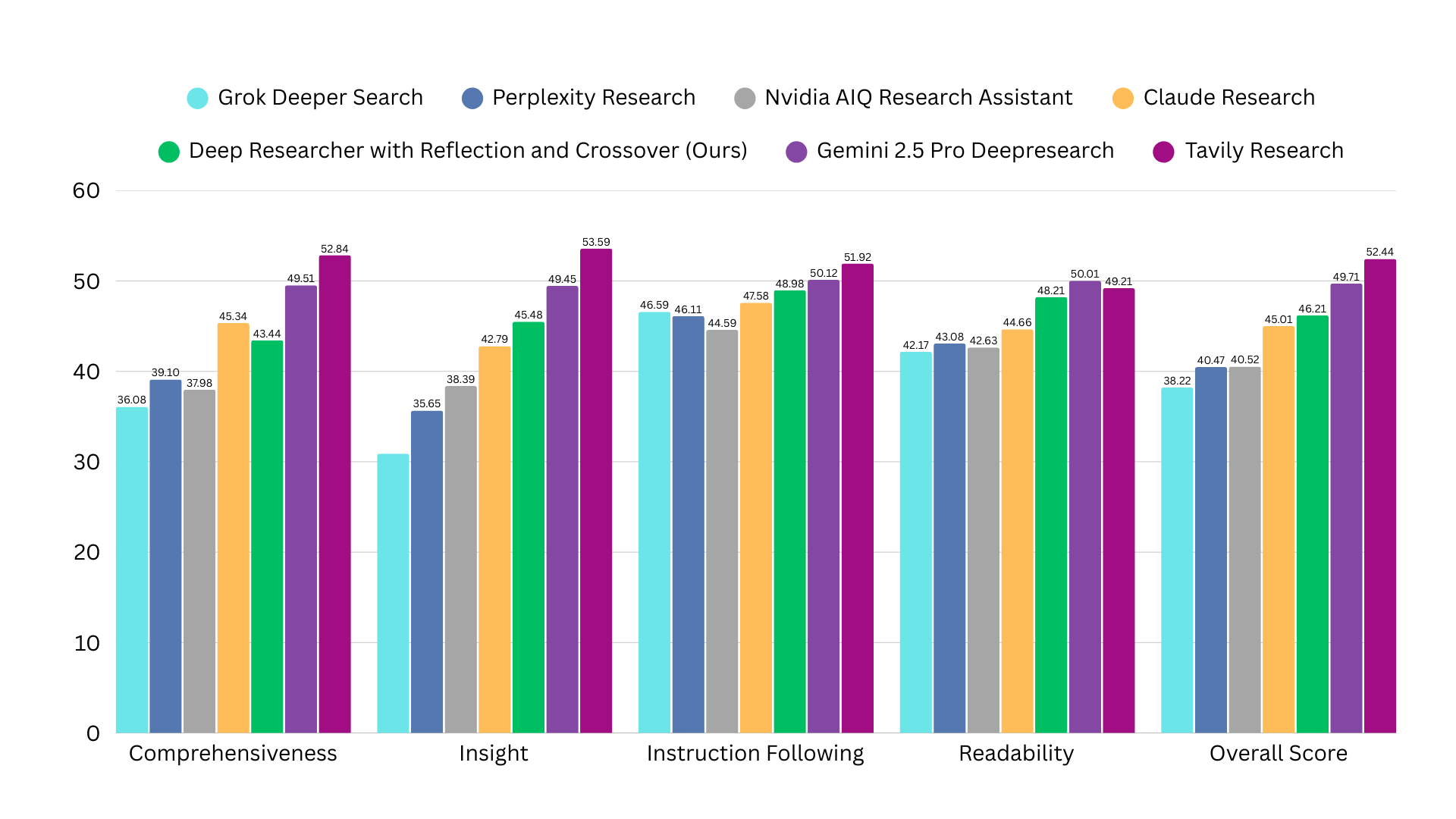}
    \caption{Comparison of Deep Researchers on 4 key dimensions of RACE framework}
    \label{fig:placeholder}
\end{figure}

\section{Introduction}

We demonstrate a Deep Researcher architecture which utilizes \textbf{Research Plan Reflection} to perform continuous plan refinement (if required) and \textbf{Candidates Crossover} allowing for the sampling of multiple answers using varied candidate’s model parameters (e.g., temperature, top\_k) to explore a larger search space. At any particular instance of time the deep researcher stores the context of the previous research done and revisits them to decide upon:

\begin{enumerate}
    \item The next (potentially un-explored) area to be researched.
    \item Refining the Research Plan if needed.
    \item Determining the percentage Research Progress.
\end{enumerate}

We continue the research process until we have hit a satisfactory threshold of research progress or we have exhausted the maximum retries. The Deep Researcher has an LLM-as-a-judge which analyzes the research performed and decides on the percentage of the research progress. If the researcher crosses the threshold of \textbf{90\%} progress, the research process is halted and a research report is generated.

We generate a research report in a single-shot by an LLM Agent acting as a report writer. The Report Writer Agent has access to the entire research context on the topic and utilises it to generate a Research Report in a single shot. Unlike Google’s TTD-DR (Test-Time Diffusion) \cite{testtimediffusiondeepresearch} which performs \textbf{Report-level Denoising} inspired by the sampling process in \textbf{Diffusion} models to where they continuously refine the noisy generated initial report iteratively.

\section{Sequential Refinement approach vs Parallel Scaling}

The development of Deep Research Agents (DRAs) has seen the emergence of two primary paradigms for handling complex, multi-faceted research tasks: Parallel Scaling and Sequential Refinement.

\begin{enumerate}
    \item \textbf{Parallel Scaling - Efficiency and Its Limitations}: Parallel scaling, as implemented in architectures like GPT Researcher \cite{gptresearcher} and our previous work, Static-DRA \cite{staticdrapaper}, focuses on decomposing a research topic into multiple independent sub-topics. These sub-topics are then investigated concurrently by parallel execution agents. While this approach offers significant advantages in terms of reduced latency and stable performance through horizontal scaling, it often suffers from a "\textbf{siloed knowledge}" problem. Because each agent operates within the vacuum of its specific sub-task, the system lacks a holistic "Global Context". This isolation makes it difficult for the model to recognize overlapping information, avoid redundant search queries, or make intelligent, real-time modifications to the research plan based on discoveries made in other branches.

    \item \textbf{Sequential Refinement - Global Context and Dynamic Adaptation}: In contrast, the Sequential Refinement approach leverages the iterative nature of the research process. Google’s \textbf{TTD-DR} (Test-Time Diffusion) \cite{testtimediffusiondeepresearch} architecture exemplifies this by performing "Report-level Denoising," where an initial draft is continuously refined through sequential iterations inspired by diffusion models. Our Deep Researcher advances this paradigm by shifting the focus from report refinement to \textbf{Sequential Research Plan Refinement}. In this model, the agent maintains a centralized \textbf{Global Research Context} - a comprehensive memory of every search trajectory and artifact gathered. By building each research chain explicitly upon previous attempts, the agent can "look back" at its progress and reason about which areas remain unexplored. This allows for dynamic plan refinement, enabling the agent to pivot its strategy at runtime, add unforeseen sub-topics, or terminate redundant paths.
\end{enumerate}

The superiority of sequential scaling is supported by recent findings in "The Sequential Edge" (Chopra 2025) \cite{sequentialedgepaper} paper, which demonstrates that sequential scaling consistently outperforms the parallel self-consistency paradigm in \textbf{95.6\%} of configurations, with accuracy gains of up to \textbf{46.7\%}. This is attributed to the model's ability to reason with a fuller, more integrated context rather than disparate fragments.

By adopting this sequential approach, our Deep Researcher achieved a score of \textbf{46.21} on the DeepResearch Bench \cite{deepresearchbenchcomprehensivebenchmark}, outperforming leading deep research agents such as \textbf{Claude} Researcher \cite{clauderesearcher}, \textbf{Perplexity} Research \cite{perplexitydeepresearch}, \textbf{Grok} Deeper Search \cite{grokdeepresearch} and many others in the leaderboard \cite{deepresearchbenchleaderboard}. Our architecture ensures that the final \textbf{One-Shot Report Generation} is informed by a unified narrative and high fact density, producing the depth required for PhD-level research.

\section{Deep Researcher Design}

\subsection{High Level Design}
The high level design of the Deep Researcher includes multiple modules working together to carry out the deep research on a given topic. The design is demonstrated in Figure 2.

\begin{figure}[H]
    \centering
    \includegraphics[width=1.0\linewidth]{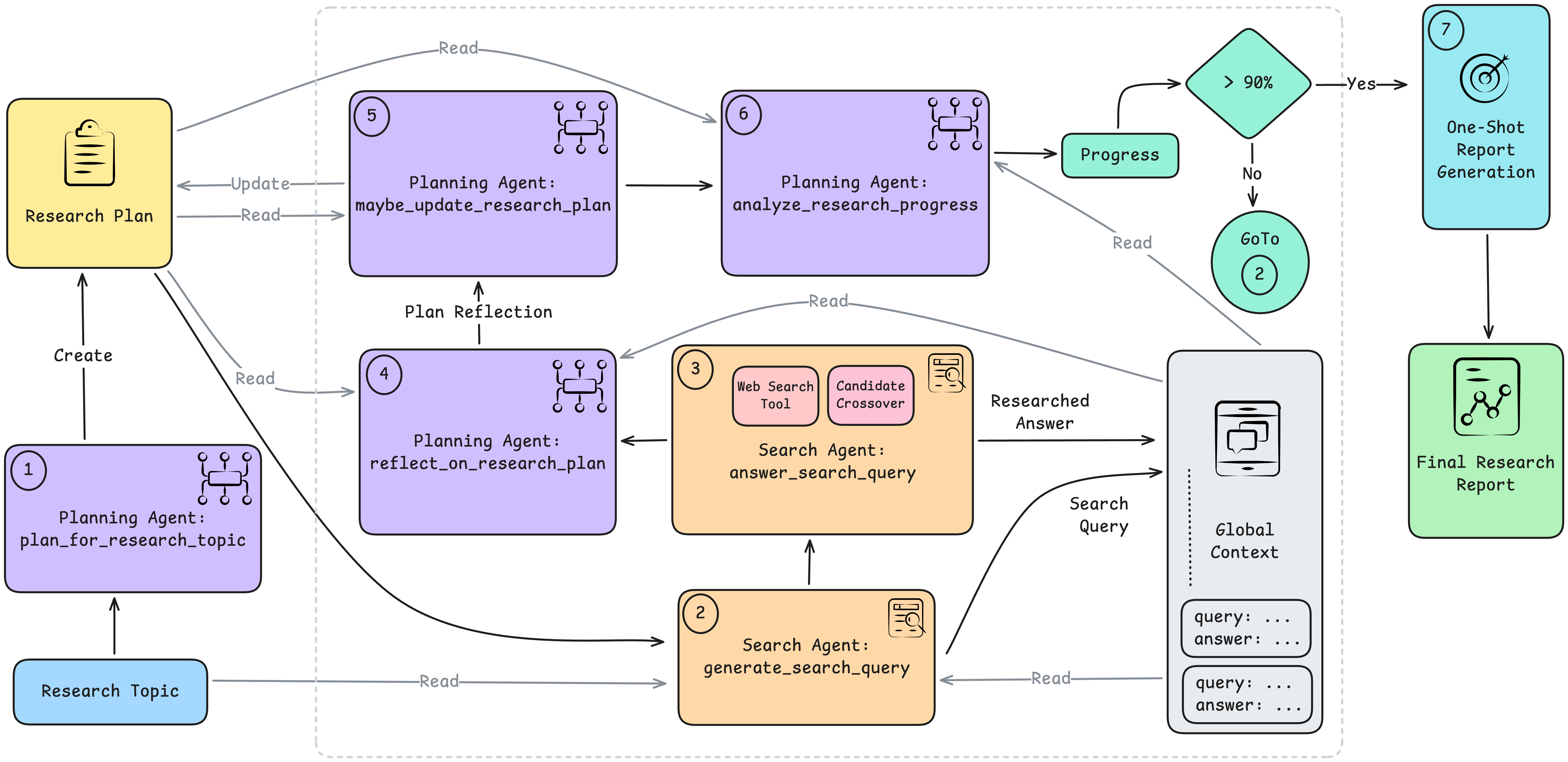}
    \caption{Deep Researcher - High Level Design}
    \label{fig:placeholder}
\end{figure}

The research methodology is structured as a series of sequential iterations, wherein each successive phase leverages findings from previous cycles to facilitate informed decision-making regarding targeted research areas and necessary plan refinements. The summary of the research process is demonstrated in the steps mentioned below.

\begin{enumerate}
    \item \textbf{Step 1 - Research Plan Curation}: The research topic is provided to the \textbf{Planning agent} that curates a research plan for the provided topic. The plan comprises detailed steps to take in order to carry out the research.
    \item \textbf{Step 2 - Generate Search Query}: The curated plan is read by the \textbf{Search agent} that generates a search query. The agent also reads the global context to understand what all has been already researched and intelligently curates a search query.
    \item \textbf{Step 3 - Answer Search Query}: The search query from the previous step is answered by the Search Agent. At this step the agent utilises a \textbf{Web Search} tool to gather recent events and updates regarding the query. The agent also incorporates the \textbf{Candidate Crossover} algorithm to improve the answer generated for the query. The search query and the answer is then added to the global context.
    \item \textbf{Step 4 - Research Plan Reflection}: The Planning agent reads the current research plan and the global context to decide whether to update the currently followed research plan or not. The agent also decides on what changes to make in the plan if at all needed.
    \item \textbf{Step 5 - Research Plan Update (maybe)}: The Planning agent takes on the plan reflection input from the previous step and makes the necessary updates in the Research Plan if suggested in the previous step. If there’s no change needed, the existing plan is followed.
    \item \textbf{Step 6 - Analyze Research Progress}: The Planning agent reads the research plan and the global context to analyze the current state of the research progress. If the research progress has crossed the \textbf{90\%} threshold benchmark, then the research process is ended. Otherwise the research loop is continued again from \textbf{Step 2}.
    \item \textbf{Step 7 - One Shot Report generation}: Once the research loop ends, we perform one-shot report generation by an LLM agent acting as a report writer. The agent is provided with the current research plan and the global context to write the research report in one go.
\end{enumerate}

The subsequent sections provide a comprehensive and detailed examination of the aforementioned procedural stages.

\subsection{Candidate Crossover algorithm}

We implement a \textbf{Candidate Crossover} algorithm that is integrated into \textbf{Step 3}, the phase in which the Search Agent conducts research for a specified query. This algorithm enhances the agent's efficiency by deploying multiple candidates to investigate the same query in parallel. Upon completion of their respective investigations, the findings are synthesized through a crossover process to generate a comprehensive and finalized research response. The algorithm is demonstrated in Figure 3.

\begin{figure}[H]
    \centering
    \includegraphics[width=0.5\linewidth]{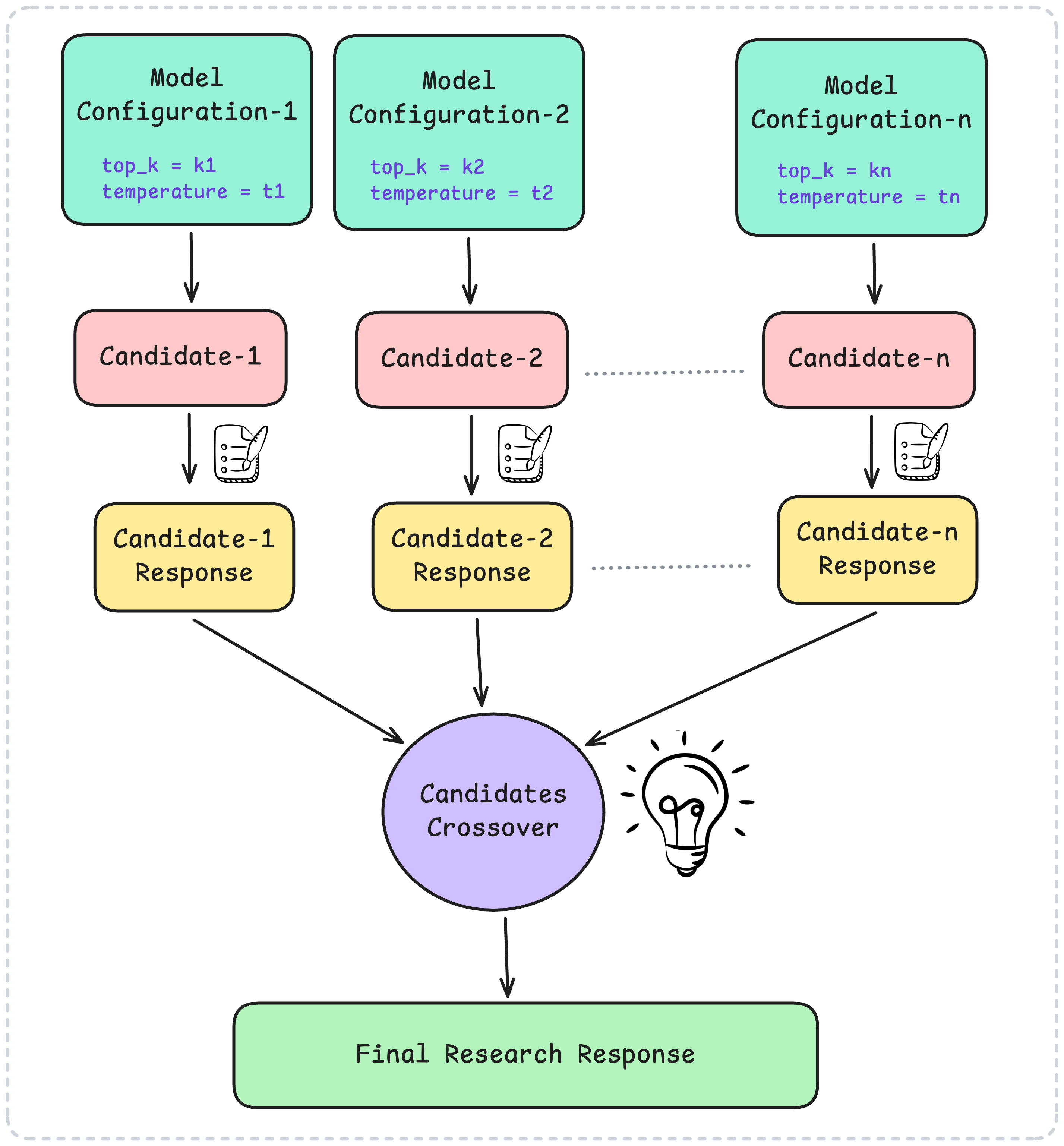}
    \caption{Deep Researcher - Candidate Crossover algorithm}
    \label{fig:placeholder}
\end{figure}

Our Candidate Crossover algorithm is inspired by the \textbf{Self-Evolution} algorithm introduced in Google’s Deep Researcher with Test Time Diffusion (\textbf{TTD-DR}) \cite{testtimediffusiondeepresearch} paper. Each candidate is a unit LLM agent with varied configuration settings. To facilitate the exploration of a large search space during inference, we initialize \textbf{n} candidates (in this paper all research topics were evaluated on \textbf{n=3} candidates), each with access to a unit LLM Agent having \textbf{n} different configurations of \textbf{temperature} and \textbf{top\_k} parameters respectively. By providing each Candidate with varied parameters, we allow each of them to attend to a different space at inference time. These candidates are provided with a search query and additional artifacts obtained from the Web Search tool, and are tasked to generate concise answers retaining all facts and numbers. We use \textbf{Tavily} \cite{tavilywebsearch} for web search and aim to receive \textbf{top 5} search results for a topic from the web. We also make sure to have only relevant web search results with us for writing a report for the research topic. The Tavily Web Search tool \cite{tavilydocs} provides a \textbf{score} field for every search result returned which defines “the relevance score of the search result”. We have a threshold score value set to \textbf{30\%} which filters out any search result whose score is less than the threshold score.

Later during the Cross-over, we combine the information by merging the answers of all the candidates, consolidating the best information from their respective evolutionary paths to curate a final research response and produce superior context for the main report generation process.

TTD-DR’s Self Evolution algorithm can be summarized in the following steps:

\begin{itemize}
    \item Step 1 - Initial States: LLM Agent units generate diverse output variants (e.g., search query answers) by sampling with varied parameters like temperature and top\_k to broaden the search space.
    \item Step 2 - Environmental Feedback: An LLM-as-a-judge uses auto-raters to evaluate variants on metrics like Helpfulness and provides textual critiques for improvement.
    \item Step 3 - Revision Step: Variants are iteratively revised based on scores and feedback until stopping criteria are met.
    \item Step 4 - Cross-over: Multiple revised variants are merged into a single high-quality output, consolidating the best information for the final report.
\end{itemize}

We did not include the \textbf{Environmental Feedback} (Step 2) and the \textbf{Revision Steps} (Step 3) present in the algorithm to reduce the latency of the Report Generation process and inference time complexity.

\subsection{Agent’s Memory: Global Research Context}

The \textbf{Global Research Context} serves as the centralized memory repository for the Deep Researcher, enabling a more cohesive sequential refinement model. This module stores the comprehensive history of the research process, including:

\begin{enumerate}
    \item \textbf{Search Trajectories}: Maintains a detailed log of every search \textbf{query} generated by the Search agent and the corresponding \textbf{answers} produced by the Search Agent with Candidate Crossover algorithm.
    \item \textbf{Contextual Artifacts}: Houses raw data, facts, and numbers gathered from \textbf{Web Search} tools, ensuring that the final report writer has access to the primary evidence discovered during the loop.
\end{enumerate}

\begin{enumerate}
    \item By maintaining this global state, the system provides the model with the "global context" necessary to reason across previously explored areas. This prevents the Search agent from drafting redundant search queries and allows the Planning agent to intelligently determine the percentage of research progress based on the totality of information gathered. The Global Research Context is particularly vital during \textbf{Step 4 (Research Plan Reflection)} and \textbf{Step 5 (Research Plan Update)}. By accessing this centralized memory, the Planning agent can perform a methodical process of reasoning that synthesizes low-level search results into higher-level insights. Specifically, the importance of the Global Context in these stages includes:

    \item \textbf{Avoiding Redundancy}: The Planning agent reviews the existing search trajectories to ensure that subsequent plan updates do not repeat previously explored queries, optimizing the efficiency of the research loop.

    \item \textbf{Dynamic Plan Refinement}: Access to the full research history (global context) enables the agent to reason about current progress and make intelligent, real-time modifications to the plan based on evidence found, rather than adhering to a rigid, pre-defined structure.

    \item \textbf{Informed Decision-Making}: The model uses the "global context" to decide which areas remain unexplored, ensuring that the updated research plan targets the most relevant and high-impact information gaps.
\end{enumerate}

Unlike parallel architectures that isolate sub-topic research presented in Static DRA [link] and GPT Researcher (link), the global research context ensures that \textbf{Step 7} (One-Shot Report Generation) is informed by a holistic understanding of the research topic, leading to more insightful and integrated final reports.

\subsection{Sequential Research Plan Refinement via Reflection}

The \textbf{Sequential Research Plan Refinement via Reflection} module is the core mechanism that enables our Deep Researcher to adapt its investigative strategy dynamically. Unlike static research architectures that follow a rigid, pre-defined path, this module empowers the Planning agent to evaluate its current progress and pivot based on the information discovered.

The refinement process is executed in two distinct phases:

\begin{enumerate}
    \item \textbf{Reflection Phase (Step 4)}: The Planning agent performs a critical review of the existing research plan against the \textbf{Global Research Context}. It assesses whether the current search results satisfy the initial research goals or if new, unforeseen sub-topics have emerged that require deeper investigation. This "look back" capability allows the agent to identify gaps in knowledge that a parallel approach might overlook.
    \item \textbf{Update Phase (Step 5)}: If the reflection phase identifies a need for adjustment, the Planning agent modifies the research plan at runtime. These modifications may include adding new research steps, re-prioritizing existing tasks, or terminating paths that have proven to be redundant.
\end{enumerate}

This sequential approach leverages findings from previous cycles to facilitate informed decision-making. By building each research chain upon the previous attempt, we align with findings from the \textbf{Sequential Edge} \cite{sequentialedgepaper} paper, which suggests that sequential scaling consistently outperforms parallel self-consistency by allowing models to reason with fuller, more integrated context. This ensures that the research trajectory remains efficient, avoiding the "siloed" knowledge problem common in parallel scaling architectures like \textbf{GPT Researcher} \cite{gptresearcher} or \textbf{Static-DRA} \cite{staticdrapaper}.

\subsection{One Shot Report Generation}

The \textbf{One Shot Report Generation} module (Step 7) serves as the final synthesis stage of the research process. Unlike architectures such as Google’s \textbf{TTD-DR} (Test Time Diffusion - Deep Research) \cite{testtimediffusiondeepresearch}, which utilize a "Report-level Denoising" process to iteratively refine a noisy initial draft through multiple diffusion-inspired steps, our system employs a single, comprehensive generation phase.

In this stage, a specialized LLM agent, designated as the \textbf{Report Writer}, is granted full access to the \textbf{Global Research Context} and the final, refined \textbf{Research Plan}. This access ensures that the agent can draw upon the entire trajectory of search queries, synthesized answers from the \textbf{Candidate Crossover} algorithm, and raw contextual artifacts such as facts and figures gathered during the sequential iterations. By processing this holistic dataset in a single inference pass, the Report Writer can:

\begin{enumerate}
    \item \textbf{Integrate Complex Information}: Synthesize findings from disparate research branches into a cohesive narrative without the "siloed" knowledge gaps common in parallel scaling architectures.
    \item \textbf{Maintain Narrative Consistency}: Ensure a unified tone and logical flow throughout the document, as the entire report is generated with the same global perspective.
    \item \textbf{Ensure Fact Density}: Utilize the centralized memory to include specific numbers, dates, and evidence discovered during the search phases, producing a detailed report suitable for PhD-level research topics.
\end{enumerate}

This approach prioritizes computational efficiency and reduced latency by avoiding the multiple refinement cycles, while still maintaining high output quality by leveraging the high-fidelity context built during the sequential reflection phases.

\section{Evaluation}

Our Deep Researcher is evaluated against the globally recognized \textbf{DeepResearch Bench} \cite{deepresearchbenchpage}. As the primary benchmark for general-purpose Deep Research Agents (DRAs), it comprises \textbf{100} doctoral-level research tasks across \textbf{22} distinct fields. This benchmark is specifically designed to assess general-purpose Deep Research Agents (DRAs). Furthermore, it implements two sophisticated evaluation frameworks to assess performance:

\begin{itemize}
    \item \textbf{RACE} (Reference-based Adaptive Criteria-driven Evaluation): This framework evaluates the qualitative merits of the final research report.
    \item \textbf{FACT} (Framework for Factual Abundance and Citation Trustworthiness): This framework assesses the agent's proficiency in data retrieval and the accuracy of its citations.
\end{itemize}

Figure 4 illustrates the allocation of 100 doctoral-level research tasks among 22 distinct academic fields. These tasks are conducted in two languages: English and Chinese. The corresponding distribution of task counts by language is also presented in Figure 4.

\begin{figure}[H]
    \centering
    \includegraphics[width=1.0\linewidth]{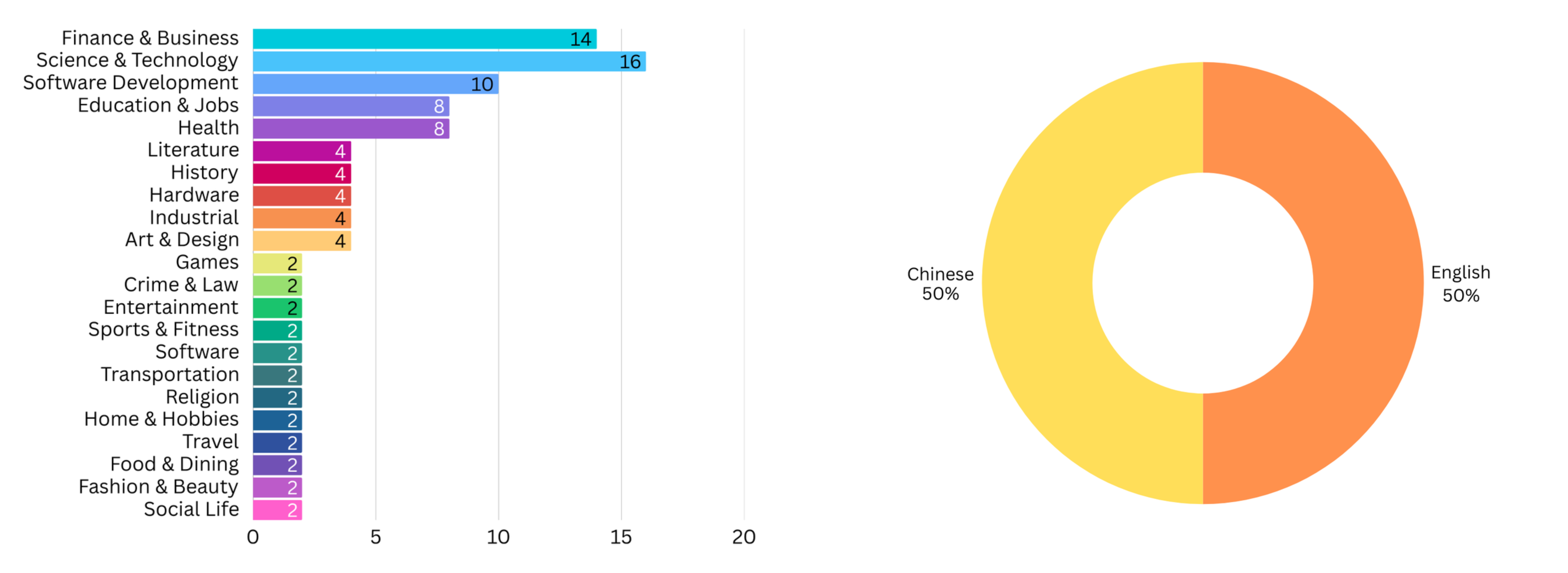}
    \caption{Allocation of tasks among fields and Distribution of task counts by language}
    \label{fig:placeholder}
\end{figure}

Our Deep Researcher underwent rigorous evaluation using the \textbf{RACE} (Reference-based Adaptive Criteria-driven Evaluation) framework, a core component of the \textbf{DeepResearch Bench}. RACE evaluates report generation quality through a sophisticated multi-step process:

\begin{enumerate}
    \item \textbf{Dynamic Criteria Generation}: Automatically generates task-specific evaluation criteria across four key dimensions:
    \begin{enumerate}
        \item \textbf{Comprehensiveness}: Coverage breadth and depth of the research topic
        \item \textbf{Insight/Depth}: Quality of analysis and insight generation
        \item \textbf{Instruction-Following}: Adherence to specific task requirements
        \item \textbf{Readability}: Clarity, organization, and presentation quality
    \end{enumerate}
    \item \textbf{Reference-Based Scoring}: Compares target reports against high-quality reference reports to ensure discriminative evaluation
    \item \textbf{Weighted Assessment}: Uses dynamic weights adapted to each task's specific requirements
\end{enumerate}

Results indicate that our architecture produces a competitive score, performing strongly against other leading deep research agents currently featured on the benchmark leaderboard \cite{deepresearchbenchleaderboard}. Our Deep Researcher established a superior position on the leaderboard, surpassing \textbf{Claude Researcher} \cite{clauderesearcher} (Overall score: \textbf{45}), \textbf{Nvidia AIQ Research Assistant} \cite{nvidiaaiqdeepresearch} (Overall score: \textbf{40.52}), \textbf{Perplexity Research} \cite{perplexitydeepresearch} (Overall score: \textbf{40.46}), and \textbf{Grok Deep Search} \cite{grokdeepresearch} (Overall score: \textbf{38.22}). The detailed comparison of our Deep Researchers on the above mentioned 4 key dimensions of the RACE framework along with the Overall Score is shown in Figure 1. Our Deep Researcher achieved a score of \textbf{48.21} on the readability metric, which represents a margin of 1 point below the state-of-the-art (SOTA) \textbf{Tavily Research} and \textbf{1.79} points below the \textbf{Gemini 2.5 Pro Deep Researcher}.

Figure 5 provides a comparative analysis of our Deep Researcher across the \textbf{four} previously defined dimensions of the \textbf{RACE} framework, including the \textbf{overall score }for each respective language. It was observed that the Deep Researcher attained a superior performance score on tasks conducted in the \textbf{Chinese} language relative to those performed in \textbf{English}.

\begin{figure}[H]
    \centering
    \includegraphics[width=1.0\linewidth]{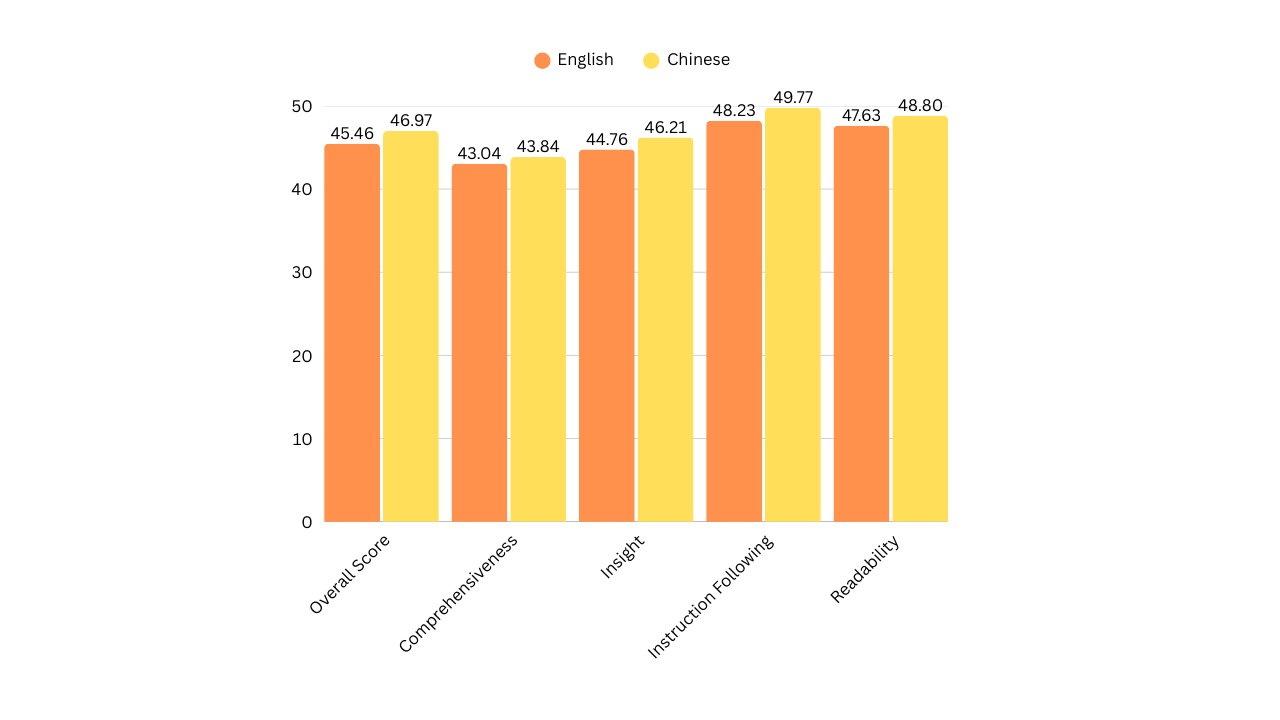}
    \caption{Comparison of Language tasks on 4 key dimensions of RACE framework}
    \label{fig:placeholder}
\end{figure}

The performance of the proposed Deep Researcher across \textbf{22} distinct academic fields is illustrated in Figure 6. The figure delineates four statistical metrics corresponding to the four key dimensions of the RACE evaluation framework.

\begin{figure}[H]
    \centering
    \includegraphics[width=1.0\linewidth]{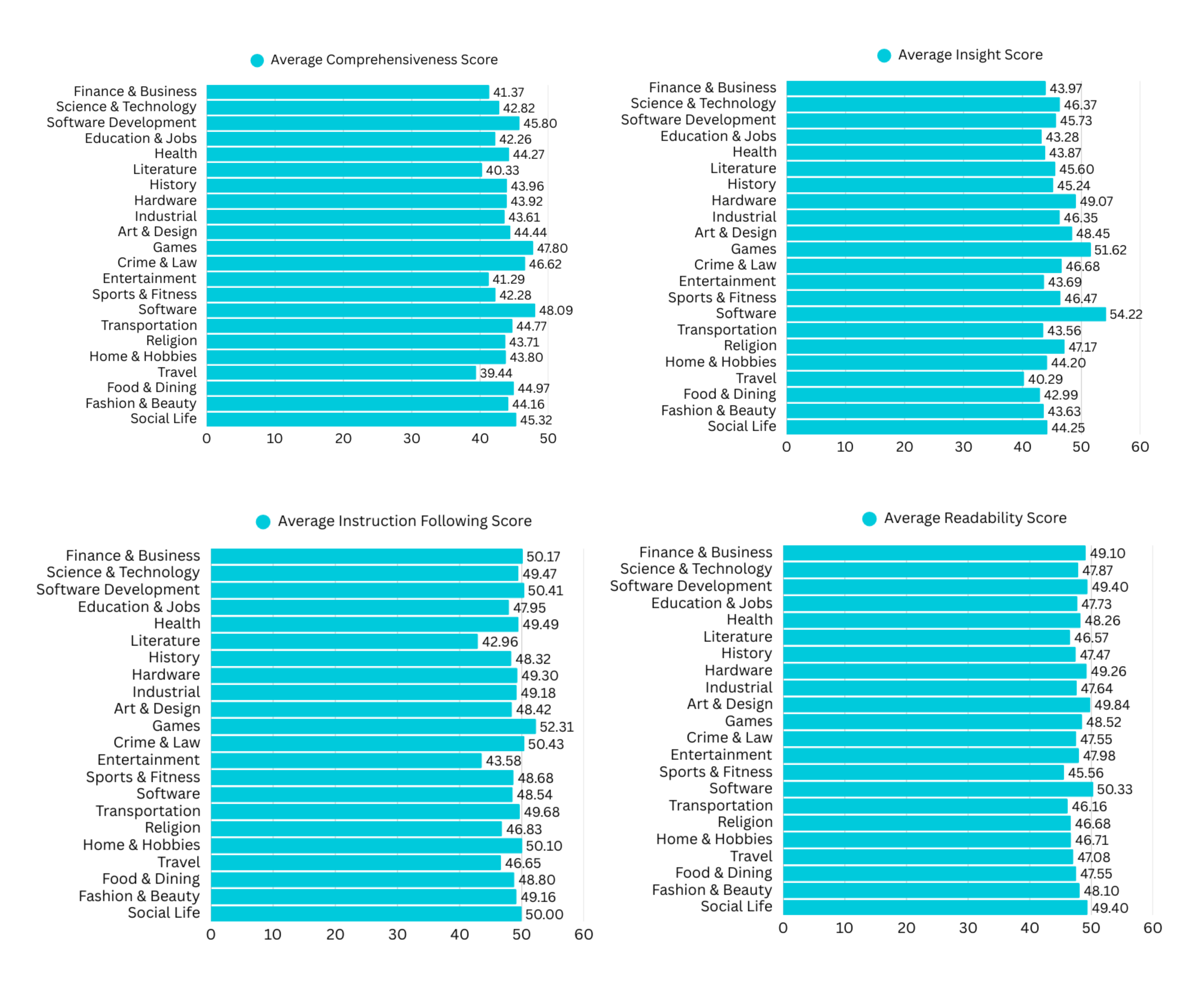}
    \caption{Deep Researcher performance across 22 distinct academic fields evaluated on 4 dimensions RACE framework}
    \label{fig:placeholder}
\end{figure}

Table 1 presents a comparative analysis of our Deep Researcher's performance against other leading deep research agents listed on the DeepResearch Bench benchmark leaderboard evaluated on the RACE framework. Additionally, Table 2 details the Deep Researcher's overall performance scores across 22 distinct academic disciplines.

\begin{table}[H] % [h] tries to place the table 'here' approximately
    \centering % Centers the table on the page
    % CAPTION GOES HERE
    \caption{Our Deep Researcher performance analysis against competitive deep research agents} 
    \label{tab:my_table} % Optional: Used for referencing later
    
    \begin{tabular}{| p{4.5cm} | p{1.5cm} | p{3cm} |  p{1.2cm} |  p{2cm} |  p{2cm} |} 
     \hline
     \textbf{Model} & 
     \textbf{Overall} & 
     \textbf{Comprehensive- ness} &
     \textbf{Insight} &
     \textbf{Instruction Following} &
     \textbf{Readability} \\ 
     \hline
     tavily-research \cite{tavilyresearch} & 52.44 & 52.84 & 53.59 & 51.92 & 49.21 \\
     \hline
     gemini-2.5-pro-deepresearch \cite{geminideepresearch} & 49.71 & 49.51 & 49.45 & 50.12 & 50 \\
     \hline
     openai-deep-research \cite{openaideepresearch} & 46.45 & 46.46 & 43.73 & 49.39 & 47.22 \\
     \hline
     deepresearcher-reflect-evolve (ours) & 46.21 & 43.44 & 45.48 & 48.99 & 48.21 \\
     \hline
     claude-research \cite{clauderesearcher} & 45 & 45.34 & 42.79 & 47.58 & 44.66 \\
     \hline
     nvidia-aiq-research-assistant \cite{nvidiaaiqdeepresearch} & 40.52 & 37.98 & 38.39 & 44.59 & 42.63 \\
     \hline
     perplexity-research \cite{perplexitydeepresearch} & 40.46 & 39.1 & 35.65 & 46.11 & 43.08 \\
     \hline
     grok-deeper-search \cite{grokdeepresearch} & 38.22 & 36.08 & 30.89 & 46.59 & 42.17 \\
     \hline
    \end{tabular}
\end{table}

\begin{table}[H] % [h] tries to place the table 'here' approximately
    \centering % Centers the table on the page
    % CAPTION GOES HERE
    \caption{Our Deep Researcher performance analysis across 22 distinct academic disciplines}
    \label{tab:my_table} % Optional: Used for referencing later
    
    \begin{tabular}{| p{3.5cm} | p{1.5cm} | p{3cm} |  p{1.5cm} |  p{2cm} |  p{2cm} |} 
     \hline
     \textbf{Academic Disciplines (Topics)} & 
     \textbf{Overall} & 
     \textbf{Comprehensive- ness} &
     \textbf{Insight} &
     \textbf{Instruction Following} &
     \textbf{Readability} \\ 
     \hline
     Finance \& Business & 45.70 & 41.36 & 43.96 & 50.16 & 49.09 \\
     \hline
     Science \& Technology & 46.39 & 42.81 & 46.37 & 49.46 & 47.86 \\
     \hline
     SoftwareDevelopment & 47.40 & 45.79 & 45.73 & 50.41 & 49.40 \\
     \hline
     Education \& Jobs & 44.86 & 42.26 & 43.28 & 47.94 & 47.73 \\
     \hline
     Health & 45.95 & 44.27 & 43.87 & 49.48 & 48.26 \\
     \hline
     Literature & 43.85 & 40.32 & 45.59 & 42.96 & 46.57 \\
     \hline
     History & 46.09 & 43.96 & 45.23 & 48.31 & 47.46 \\
     \hline
     Hardware & 47.60 & 43.92 & 49.06 & 49.29 & 49.25 \\
     \hline
     Industrial & 46.37 & 43.61 & 46.34 & 49.18 & 47.64 \\
     \hline
     Art \& Design & 47.50 & 44.44 & 48.44 & 48.42 & 49.83 \\
     \hline
     Games & 50.36 & 47.80 & 51.62 & 52.31 & 48.51 \\
     \hline
     Crime \& Law & 47.59 & 46.62 & 46.67 & 50.43 & 47.54 \\
     \hline
     Entertainment & 43.20 & 41.29 & 43.68 & 43.58 & 47.98 \\
     \hline
     Sports \& Fitness & 45.62 & 42.28 & 46.46 & 48.68 & 45.56 \\
     \hline
     Software & 50.78 & 48.08 & 54.21 & 48.54 & 50.33 \\
     \hline
     Transportation & 46.02 & 44.76 & 43.56 & 49.68 & 46.15 \\
     \hline
     Religion & 45.95 & 43.71 & 47.16 & 46.83 & 46.67 \\
     \hline
     Home \& Hobbies & 45.94 & 43.80 & 44.20 & 50.10 & 46.70 \\
     \hline
     Travel & 42.43 & 39.44 & 40.28 & 46.64 & 47.07 \\
     \hline
     Food \& Dining & 46.09 & 44.97 & 42.98 & 48.79 & 47.55 \\
     \hline
     Fashion \& Beauty & 45.76 & 44.16 & 43.63 & 49.15 & 48.10 \\
     \hline
     Social Life & 46.74 & 45.31 & 44.25 & 50.00 & 49.39 \\
     \hline
    \end{tabular}
\end{table}

\section{Conclusion}

This paper introduced the novel \textbf{Deep Researcher} architecture, which shifts the paradigm from latency-optimized parallel scaling to an accuracy-driven sequential refinement model. The system's core innovations are the \textbf{Sequential Research Plan Refinement via Reflection} and the \textbf{Candidates Crossover} algorithm. Sequential refinement enables the agent to maintain a centralized \textbf{Global Research} Context, allowing it to dynamically adapt its research plan, avoid redundant searches, and overcome the "siloed knowledge" problem inherent in parallel architectures like Static-DRA and GPT Researcher. The Candidates Crossover algorithm further optimized search efficiency by deploying multiple LLM agents with varied parameters to explore a larger search space, with their findings synthesized for a comprehensive final response.

The effectiveness of this approach was demonstrated through rigorous evaluation on the \textbf{DeepResearch Bench}, a global benchmark of 100 doctoral-level research tasks. Powered by the \textbf{gemini-2.5-pro} model, our Deep Researcher achieved a superior overall score of \textbf{46.21}, significantly surpassing several leading deep research agents. These results reinforce the critical finding that sequential scaling consistently outperforms the parallel self-consistency paradigm, validating the system's ability to generate highly detailed, fact-dense reports suitable for PhD-level research using a \textbf{One Shot Report Generation} process that maintains computational efficiency.

\bibliographystyle{plain} % or unsrt, ieeetr
\bibliography{references}

\end{document}